\documentclass{article}
\usepackage[utf8]{inputenc}
\usepackage{comment}
\usepackage{subfig}
\usepackage{graphicx}
\usepackage{verbatim}
\usepackage{tabularx} 
\usepackage{rotating}
\usepackage{float}
\usepackage{amsmath,amsfonts,amssymb}
\usepackage{physics}
\title{A Survey on Activation Functions and their relation with Xavier and He Normal Initialization}
\author{Leonid Datta}
\date{Delft University of Technology}

\begin{document}

\maketitle
\begin{abstract}
    In artificial neural network, the activation function and the weight initialization method play important roles in training and performance of a neural network. The question arises is what properties of a function are important/necessary for being a well-performing activation function. Also, the most widely used weight initialization methods - Xavier and He normal initialization have fundamental connection with activation function. This survey discusses the important/necessary properties of activation function and the most widely used activation functions (sigmoid, tanh, ReLU, LReLU and PReLU). This survey also explores the relationship between these activation functions and the two weight initialization methods - Xavier and He normal initialization.
\end{abstract}

\section{Introduction}
Artificial intelligence has been trying to make intelligent machines for long \cite{mccarthy2004artificial}. Artificial neural networks have played important roles in artificial intelligence to achieve its goal \cite{haykin99a} \cite{Bishop:1995:NNP:525960}. When an artificial neural network is built to execute a task, it is programmed to perceive a pattern. The main task of an artificial neural network is to learn this pattern from data \cite{mccarthy2004artificial}.  

An artificial neural network is composed of large number of interconnected working units known as perceptrons or neurons. A perceptron is composed of four components: input node, weight vector, activation function and output node.  The first component of a perceptron is the input node - it receives the input vector. I assume to have an $m$ dimensional input vector  $\mathbf{x}=\begin{bmatrix}
           x_{1} \  
           x_{2} \ 
           \hdots 
           x_{m} \ 
         \end{bmatrix}$. 

The second component is the weight vector which has the same dimension as that of the input vector. Here, the weight vector is $\mathbf{w}= \begin{bmatrix}
           w_{1}  \ 
           w_{2} \ 
           \hdots 
           w_{m} \ 
         \end{bmatrix}$. 
From the input vector and the weight vector, an inner product is calculated as $\mathbf{x}^\top\mathbf{w}$. While calculating the inner product, an extra term 1 is added to the input vector with an initial weight value (here $w_0$). This is known as bias. This bias term is added as an intercept so that the inner product can be easily shifted using the weight $w_0$. After the bias has been added, the input vector and the weight vector look like  $\mathbf{x}=\begin{bmatrix}
            1   \ 
           x_{1}    \ 
           x_{2}    \ 
           \hdots \ 
           x_{m}    \ 
         \end{bmatrix}$ and $\mathbf{w}=\begin{bmatrix}
            w_{0}   \ 
           w_{1}    \ 
           w_{2}    \ 
           \hdots \
           w_{m}    \ 
         \end{bmatrix}$. Thus, $\mathbf{x}^\top\mathbf{w} =w_0 + w_1x_1 + w_2x_2 + \ldots + w_{m}x_{m}$.

The inner product $\mathbf{x}^\top\mathbf{w}$ is used as input to a function known as the activation function (here symbolized as $f$). This activation function is the third component of the perceptron. The output from the activation function is sent to the output node which is the fourth component. This is how an input vector flows from the input node to the output node. It is known as forward propagation. 

In mathematical terms, the output $y(x_1,..., x_{m})$ can be expressed as 
 \begin{equation}
    y(x_1,\ldots,x_{m}) = f(w_0 + w_1x_1 + w_2x_2 + \ldots + w_{m}x_{m})
  \end{equation}.

\begin{center}
\begin{figure}[h!]
  \includegraphics[scale=0.45]{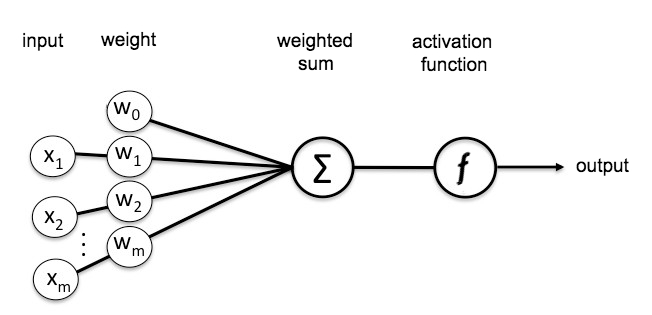}
  \caption{Structure of a single layer perceptron}
  \label{fig:nn}
\end{figure}
\end{center}

The perceptron explained above is a single layer perceptron as it has only one layer of weight vector and activation function. This structure is shown in fig \ref{fig:nn}. Neural networks typically have more than one layer. Neural networks receive the input and pass them through a series of weight vector and activation function. At the end, it is passed at the output layer. Since, the layers between the input layer and the output layer are not directly visible from outside, they are called the hidden layers. A neural network with one hidden layer is known as shallow network.

When the output is produced at the output layer, the neural network calculates the loss. Loss quantifies the difference between the obtained output and the desired output. Let $L$ be the loss. The goal of the neural network is to update the weight vectors of the perceptron to reduce the loss $L$.
To perform this task, gradient descent is used. In gradient descent, the weights are updated by moving them to the opposite direction of the gradient of the loss with respect to weight. The gradient of the loss with respect to the weights $\frac{\partial L}{\partial \mathbf{w}}$ is expressed as \begin{equation} \frac{\partial L}{\partial \mathbf{w}}=\frac{\partial L}{\partial y}\cdot\frac{\partial y}{\partial \mathbf{w}}=\frac{\partial L}{\partial y}\cdot\frac{\partial y}{\partial z}\cdot \frac{\partial z}{\partial \mathbf{w}} \end{equation} where $z$ is the inner product $z=\mathbf{x}^\top\mathbf{w}$ and $y=f(z)$. This is known as the chain rule. This update is done in the opposite direction of the forward propagation- starting from the last layer and then gradually moving to the first layer. This process is known as backward propagation or backpropagation. One forward propagation and one backward propagation of all the training examples or training data is termed as an epoch.

Since the weight vector gets updated during training, it needs to be assigned initial values before the training starts. Assigning the initial values to the weight vector is known as weight initialization. Once the weight vector is updated, the input is again passed through the network in forward direction to generate the output and calculate the loss. This process continues till the loss reaches a satisfactory minimum value. A network is said to converge when the loss achieves the satisfactory minimum value and this process is called the training of a neural network. When an artificial neural network aims to perform a task, the neural network is first trained.

One of the most interesting characteristics of artificial neural network is the possibility to adapt its behavior with the changing characteristics of the type of task. The activation function has an important role in this behaviour adaptation and learning \cite{agostinelli2014learning}. The choice of the activation function at different layers in a neural network is important as it decides how the data will be presented to the next layer. Also, it controls how much bounded or not bounded (that is, the range of the data) the data will be to the next layer or the output layer of the network. When the network is sufficiently deep, learning the pattern becomes less difficult for common nonlinear functions \cite{hornik1989multilayer}  \cite{cho2009kernel}. But when the network is not complex and deep, the choice of activation function has more significant effect on the learning pattern and the performance of the neural network than that of the complex and deep networks \cite{agostinelli2014learning}.

Weight initialization takes an important role in the speed of training a neural network \cite{bottou1988reconnaissance}. More precisely it controls the speed of the backpropagation because the weights are updated during backpropagation \cite{drago1992statistically}. If the weight initialization is not proper, it can lead to poor update of weight so that the weights will get saturated at early stage of training\cite{yam2000weight}. This can cause the training process to fail. In artificial neural network,Xavier  and He normal weight initialization method have gained popularity among the different weight initialization methods since they have been proposed \cite{glorot11a} \cite{heetal}\cite{goodinit} \cite{ronneberger2015u} \cite{huang2017densely} \cite{lecun2015deep}.

\subsection{Contribution}
The main contributions of this survey are: 
\begin{enumerate}
    \item This survey discusses the necessary/ important properties an activation function is expected to have and explore why they are necessary/important.
    \item This survey discusses the four widely used functions - sigmoid, tanh, ReLU, LReLU and PReLU and why they are so widely used. It also discusses the problems faced by these activation functions and why they face the problems.
    \item This survey discusses Xavier  and He normal weight initialization method and their relation with the mentioned activation functions .   
    
\end{enumerate}

\subsection{Organisation of the report}
 
In this report, section 2 puts light on the activation function in detail. Section 2.1 talks about the necessary/important properties an activation function is expected to have. Section 2.2 talks about the main two problems (vanishing gradient problem and dead neuron problem)  faced by activation functions. Section 2.3 discusses the five different activation functions (sigmoid, tanh, ReLU, LReLU and PReLU), their properties and problems faced by them. Section 2.4 shows which properties which properties hold or do not hold by the mentioned activation functions and also the problems faced or not faced by the mentioned activation functions in a tabular format.

Section 3 focuses on the weight initialization methods. Section 3.1 discusses the Xavier initialization method and section 3.2 discusses the  He normal weight initialization method  \cite{glorot11a} \cite{heetal}. 

Section 4 gives an overview of the insights from the survey and finally the conclusions of the survey are noted in section 5.

\section{Activation function}
The activation function, also known as the transfer function, is the nonlinear function applied on the inner product $\mathbf{x}^\top\mathbf{w}$ in an artificial neural network. 

Before discussing the properties of activation function, it is important to know what the sigmoidal activation function is. A function is called sigmoidal in nature if it has all the following properties:
\begin{enumerate}
    \item it is continuous
    \item it is bounded between a finite lower and upper bound
    \item it is nonlinear
    \item its domain contains all real numbers
    \item its output value monotonically increases when the input value increases and as a result of this, it has got an `S' shaped curve.
\end{enumerate}
A feed-forward network having a single hidden layer with finite number of neurons and sigmoidal activation function  can approximate any continuous boundary or functions. This is known as the Cybenko theorem \cite{Cybenko1989}.

\subsection{Properties of activation function}
The properties activation functions are expected to have are-
\begin{enumerate}
    \item Nonlinear: There are two reasons why an activation function should be nonlinear. They are:
    \begin{enumerate}
        \item The boundaries or patterns in real-world problems cannot always be expected to be linear. A non-linear function can easily approximate a linear boundary whereas a linear function cannot approximate a non-linear boundary. Since an artificial neural network learns the pattern or boundary from data, nonlinearity in activation function is necessary so that the artificial neural network can easily learn any linear or non-linear boundary.
        \item If the activation function is linear, then a perceptron with multiple hidden layers can be easily compressed to a single layer perceptron because a linear combination of another linear combination of input vector can be simply expressed as a single linear combination of the input vector. In that case, the depth of the network will have no effect. This is another reason why activation functions are non-linear. 
    \end{enumerate}
    So, nonlinearity in activation function is necessary when the decision boundary is nonlinear in nature.  
    \item Differentiable: During backpropagation, the gradient of the loss function is calculated in gradient descent method. The gradient of the loss function with respect to weight is calculated as $\frac{\partial L}{\partial \mathbf{w}}=\frac{\partial L}{\partial y}\cdot\frac{\partial y}{\partial z}\cdot \frac{\partial z}{\partial \mathbf{w}} $ as explained in equation number (2). The term $ \frac{\partial y}{\partial z} = \frac{\partial f(z)}{\partial z}$  appears in the gradient expression. So, it is necessary that the activation function is differentiable with respect to its input. 
    
    \item Continuous: A function cannot be differentiable unless it is continuous. Differentiability is a necessary property of activation function. This makes continuity a necessary property for an activation function.
    
    \item Bounded: The input data is passed through a series of perceptrons each of which contains an activation function. As a result of this, if the function is not bounded in a range, the output value may explode. To control this explosion of values, a bounded nature of activation function is important but not necessary.  
    
    \item Zero-centered: A function is said to be zero centered when its range contains both positive and negative values. If the activation function of the network is not zero centered, $y=f(\mathbf{x}^\top\mathbf{w})$ is always positive or always negative. Thus, the output of a layer is always being moved to either the positive values or the negative values. As a result, the weight vector needs more update to be trained properly. So, the number of epochs needed for the network to get trained  increases if the activation function is not zero centered. This  is why the zero centered property is important, though it is not necessary. 
    
    \item Computational cost: The computational cost of an activation function is defined as the time required to generate the output of the activation function when  input is fed to it. Along with the computational cost of the activation function, the computational cost of the gradient is also important as the gradient is calculated during weight update in backpropagation.
    
    Gradient descent optimization itself is a very time consuming process and many iterations are needed to perform this. Therefore, the computational cost is an issue.
    In an artificial neural network, if the activation function or gradient of the activation function  has low computational cost, it requires less time to get trained. When  the activation function and the gradient of the activation function have high computational cost, it requires more time to get trained. An activation function with less computational cost of it and its' gradient is preferable as it saves time and energy. 
    \end{enumerate}
    \subsection{Problems faced by activation functions}
    The vanishing gradient problem and the dead neuron problem are the two major problems faced by the widely used activation functions. These two problems are discussed in this section.
    \subsubsection{Vanishing gradient problem  }

    When an activation function compresses a large range of input into a small output range, a large change to the input of the activation function results into a very small change to the output. This leads to a small (typically close to zero) gradient value. 
    
    During weight update, when backpropagation calculates the gradients of the network, the gradients of each layer are multiplied down from the final layer to that layer because of the chain rule. When a value close to zero is multiplied to other values close to zero several times, the value becomes closer and closer to zero. In this same way, the values of the gradients get closer and closer to 0 as the backpropagation goes deeper into the network. So, the weights get saturated and they are not updated properly. As a result, the loss stops decreasing and the network does not get trained properly. This problem is termed as the vanishing gradient problem. The neurons, of which the weights are not updated properly are called the saturated neurons. 

\subsubsection{Dead neuron problem}

    When an activation function forces a large part of the input to zero or almost zero, those corresponding neurons are inactive/dead in contributing to the final output. During weight update, there is a possibility that the weights will be updated in such a way that the weighted sum of a large part of the network will be forced to zero. A network will hardly recover from such a situation and a large portion of the input fails to contribute to the network. This leads to a problem because a large part of the input may remain completely deactivated during the network performs. These forcefully deactivated neurons are called 'dead neurons' and this problem is termed as the dead neuron problem.

\subsection{Widely used activation functions}

In this section, we will look into the most widely used activation functions and their properties.
\subsubsection{Sigmoid/ Logistic sigmoid function}

The logistic sigmoid function, sometimes referred as only `sigmoid function’, is one of the most used activation functions in artificial neural networks \cite{DBLP} . The function is defined as $$ f(x) =  \frac{\mathrm{1} }{\mathrm{1} + e^{-x}}  $$ where $x$ is the input to the activation function.

\begin{figure}[H]%
    \centering
    \subfloat 2(a)   {{\includegraphics[scale=0.3]{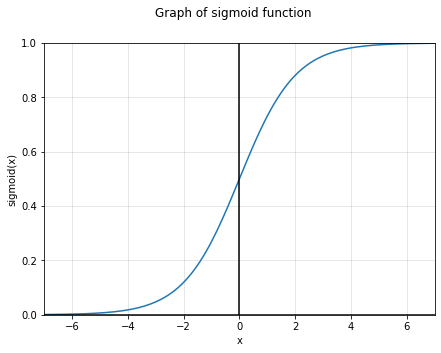} }}%
    \qquad
    \subfloat2(b){{\includegraphics[scale=0.3]{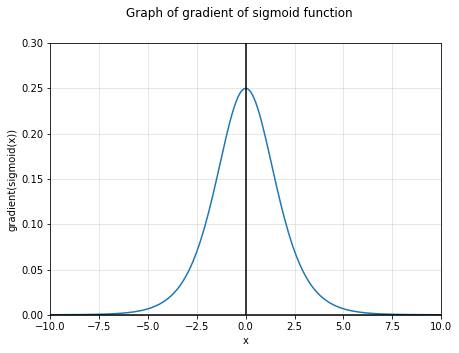} }}%
    \caption{2(a): Graph of the sigmoid function 2(b): Graph of gradient of sigmoid function}%
    \label{fig:sigg}%
\end{figure}

The sigmoid function is continuous and bounded in the range of (0,1) and differentiable. More importantly, the sigmoid function belongs to the sigmoidal group of activation functions. So, the Cybenko theorem holds true for the sigmoid function. These are why it is widely used \cite{DBLP} \cite{Hornik}. The graph of the sigmoid function and its gradient is shown in figure \ref{fig:sigg}.

The sigmoid function contains an exponential term as it can be seen from the function definition. Exponential functions have high computation cost and as a result of this, the sigmoid function has a high computational cost. Although, the function is computationally expensive, its gradient is not. Its gradient can be calculated using the formula  $f'(x)=f(x)(1-f(x))$.

The sigmoid function suffers some major drawbacks as well. The sigmoid function is bound in the range of (0,1). Hence it always produces a non-negative value as output. Thus it is not a zero-centered activation function. The sigmoid function binds a large range of input to a small range of (0,1). So, a large change to the input value leads to a small change to the output value. This results to small gradient values as well. Because of the small gradient values, it suffers the vanishing gradient problem.

To have the benefits of the sigmoid function along with zero-centered nature, the hyperbolic tangent or tanh function was introduced.

\subsubsection{Tanh function}
The hyperbolic tangent or tanh function is slightly more popular than the sigmoid function because it gives better training performance for multi-layer neural networks \cite{Karl} \cite{Neal}. The tanh function is defined as $$ f(x) =  \frac{\mathrm{1}- e^{-x} }{\mathrm{1} + e^{-x} }  $$ where $x$ is the input to the activation function.

Tanh function can be seen as a modified version of the sigmoid function because tanh can be expressed as $tanh(x)= 2sigmoid(2x)-1$. That is the reason why it has got all the properties of the sigmoid function. It is continuous, differentiable and bounded. It has got a range of (-1,1). So, it produces negative, positive and zero as outputs. Thus it is zero centered activation function and solves the ‘not a zero-centered activation function’ problem of the sigmoid function. 

\begin{figure}[H]%
    \centering
    \subfloat 3(a)   {{\includegraphics[scale=0.3]{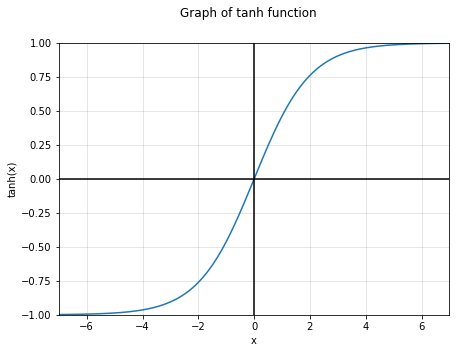} }}%
    \qquad
    \subfloat 3(b){{\includegraphics[scale=0.3]{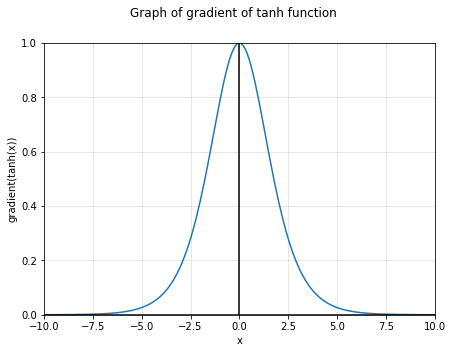} }}%
    \caption{3(a): Graph of the tanh function 3(b): Graph of gradient of tanh function}%
    \label{fig:tanhh}%
\end{figure}

The tanh function also belongs to the sigmoidal group of function and thus the Cybenko theorem holds true for tanh as well. The main advantage provided by the tanh function is that it produces zero centered output and thereby it aids the back-propagation process\cite{nwankpa}.

Tanh is computationally expensive for the same reason as that of sigmoid - it is exponential in nature. Though, the gradient calculation is not expensive. The gradient of tanh can be calculated using $f'(x)=(1-f(x)^2$). The graph of the tanh and the corresponding gradient is shown in figure \ref{fig:tanhh}. 

The tanh function, in a similar way to sigmoid, binds a large range of input to a small range of (-1,1). So, a large change to the input value leads to a small change to the output value. This results into close to zero gradient values. Because of the close to zero gradient values, tanh suffers vanishing gradient problem. The problem of vanishing gradient spurred further research in activation functions and ReLU was introduced. 

\subsubsection{ReLU function}
The ReLU or rectified linear unit function was proposed in the paper Nair et al. \cite{inproceedings}. This function is defined as $f(x)=max(0,x)$ where x is the input to the activation function. ReLU has been one of the most widely used activation functions since it has been proposed \cite{ramachandran2018searching}.

\begin{figure}[H]%
    \centering
    \subfloat 4(a)   {{\includegraphics[scale=0.3]{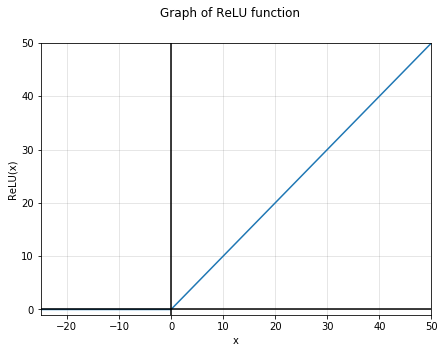} }}%
    \qquad
    \subfloat4(b){{\includegraphics[scale=0.3]{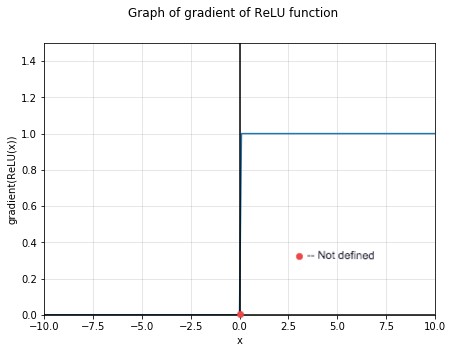} }}%
    \caption{4(a): Graph of the ReLU function 4(b): Graph of gradient of ReLU function}%
    \label{fig:relu}%
\end{figure}

The ReLU function is continuous, not-bounded and not zero-centered. At $x=0$, the left hand derivative of ReLU is zero while right hand derivative is 1. Since the left hand derivative and the right hand derivative are not equal at $x=0$, ReLU function is not differentiable at $x=0$.  ReLU has got extremely cheap computational cost as it basically pushes the negative values to zero. It also introduces sparsity to the problem in the same way \cite{ramachandran2018searching}. The paper Krizhevsky et al. demonstrated that ReLU is six times faster than sigmoid/tanh in terms of number of epochs required to train a network \cite{krizhevsky2012imagenet}. The ReLU function offers better performance and generalization in neural networks than the sigmoid and tanh activation functions and is generally used within the hidden units in the deep neural networks with another activation functions in the output layers of the network  \cite{40811} \cite{conf/icassp/DahlSH13}. 

Glorot et al. 2011 states that there is no vanishing gradient problem in the ReLU and Lau et al. 2017 quotes the same \cite{glorot11a} \cite{Lau}. Maas et al. 2013 describes that the ReLU activates values above zero and its partial derivative is 1. Since, the derivative is exactly 0 or 1 all the time, the vanishing gradient problem do not exist in network \cite{Maas13rectifiernonlinearities}. 
The graph of the ReLU function and the corresponding gradient is shown in figure \ref{fig:relu}.

Any input to the ReLU function which is less than zero generates zero as output. As a result, the parts of the input which have negative weighted sums fail to contribute to the whole process. Thus, ReLU can make a network fragile by deactivating a large part of it. Hence, it from suffers from the dead neuron which is already explained in section 2.2.2. This is also termed as dying ReLU problem and the neurons which are deactivated by ReLU are called dead neurons.

The dying ReLU problem lead forwards to a new variant of the ReLU called the Leaky ReLU or LReLU.

\subsubsection{Leaky ReLU function}
The lealy ReLU or LReLU function was introduced by Maas et al. 2013 \cite{Maas13rectifiernonlinearities}. The LReLU function is defined as 
\[
  f(x) =
  \begin{cases}
                                   0.01x  &  \text{ for $x\leq 0$} \\
                                   x & \text{ otherwise} \\
  
  \end{cases}
\] where $x$ is the input to the activation function.

\begin{figure}[H]%
    \centering
    \subfloat 5(a)   {{\includegraphics[scale=0.3]{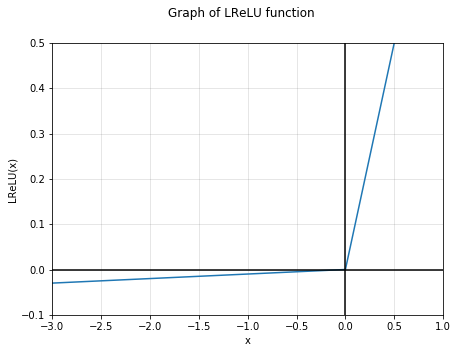} }}%
    \qquad
    \subfloat5(b){{\includegraphics[scale=0.3]{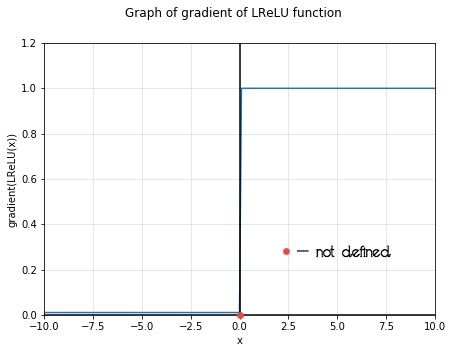} }}%
    \caption{5(a): Graph of the LReLU function 5(b): Graph of gradient of LReLU function}%
    \label{fig:lrelu}%
\end{figure}

The LReLU function is continuous and not-bounded. It is computationally very cheap and a zero-centered activation function. The LReLU function allows a small part of negative units instead of pushing them to zero like the ReLU function does. Because of this, it loses the sparse nature of the ReLU function. 

 At $x=0$, the left hand derivative of LReLU is 0.01 while right hand derivative is 1. Since the left hand derivative and the right hand derivative are not equal at $x=0$, LReLU function is not differentiable at $x=0$. In the positive part of LReLU function where gradient is always 1, there no vanishing gradient problem. But on the negative part, the gradient is always 0.01 which is close to zero.  It leads to a risk of vanishing gradient problem.

The graph of LReLU and its gradient is shown in figure \ref{fig:lrelu}.

\subsubsection{Parametric ReLU function}
The parametric ReLU or PReLU function was introduced by He et al.  \cite{heetal}. It is defined as \[
  f(x) =
  \begin{cases}
                                   ax  &  \text{ for $x\leq 0$} \\
                                   x & \text{ otherwise} \\
  
  \end{cases}
\] where $a$ is a learnable parameter and $x$ is the input to the activation function. When this $a$ is 0.01, PReLU function becomes leaky ReLU and for $a=0$ PReLU function becomes ReLU. That is why PReLU can be seen as a general representation of rectifier nonlinearities. 

\begin{figure}[H]%
    \centering
    \subfloat 6(a)   {{\includegraphics[scale=0.3]{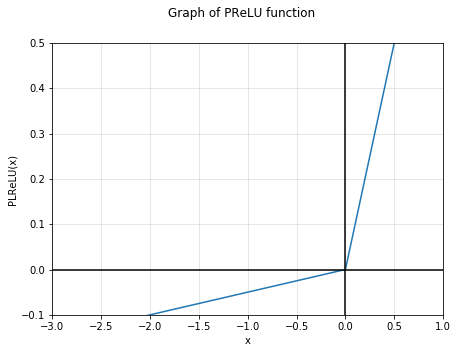} }}%
    \qquad
    \subfloat 6(b){{\includegraphics[scale=0.3]{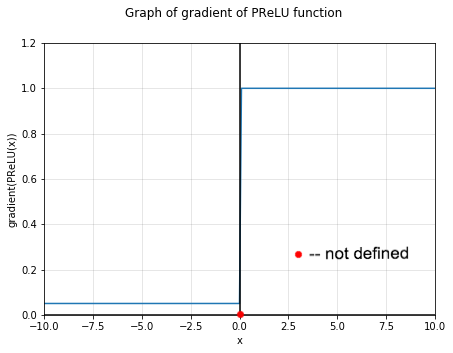} }}%
    \caption{6(a): Graph of the PReLU function 6(b): Graph of gradient of PReLU function}%
    \label{fig:prelu}%
\end{figure}

This PReLU function is continuous, not bounded and zero centered. When $x<0$, then the gradient of the function is $a$, and when $x>0$ the gradient of the function is 1. This function is not differentiable at $x=0$ because the right hand derivative and left hand derivative are not equal at $x=0$. In the positive part of PReLU function where gradient is always 1, there no vanishing gradient problem.  But on the negative part, the gradient is always $a$ which  is typically close to zero.  It leads to a risk of vanishing gradient problem. The graph of the PRelu function and its gradient is shown in figure \ref{fig:prelu}.

\subsection{Shared properties among different activation functions }
In this section, the shared properties among different activation functions are shown in tabular form in table \ref{tab:table1}, \ref{tab:table2} and \ref{tab:table3}. Table \ref{tab:table1} shows the computational cost of the functions and the same of the gradients. Table \ref{tab:table2} shows the  shared properties among the different activation functions. Table \ref{tab:table3} shows which activation function faces dead neuron problem and solves the vanishing gradient problem.
\begin{table}[H]
\centering
\caption{Calculation cost of activation functions and their gradients  }
\label{tab:table1}
\begin{tabular}{c|c|c}
Function & Computational cost & Gradient computational cost\\
\hline
Logistic sigmoid & High & Low \\
Tanh & High & Low \\
ReLU & Low & Low \\
LReLU & Low & Low \\
PReLU & Low & Low \\

\end{tabular}
\end{table}

\begin{table}[H]
\centering
\caption{Shared properties among different activation functions}
\label{tab:table2}
\begin{tabular}{c|c|c|c|c|c}
Function & Bounded? & Sigmoidal? & Continuous? & Zero-centered? & Differentiable? \\
\hline
Sigmoid & Yes & Yes & Yes & No & Yes \\
Tanh & Yes & Yes & Yes & Yes & Yes \\
ReLU & No & No & Yes & No & Yes (except at x=0) \\
LReLU & No & No & Yes & Yes & Yes(except at x=0) \\
PReLU & No & No & Yes & Yes & Yes(except at x=0) \\
\end{tabular}
\end{table}

\begin{table}[H]
\centering
\caption{Problems faced by different activation functions  }
\label{tab:table3}
\begin{tabular}{c|c|c}
Function & Faces dead neuron problem? & Faces vanishing gradient problem?\\
\hline
Logistic sigmoid & No & Yes \\
Tanh  & No & Yes \\
ReLU & Yes & No \\
LReLU & No & Partially \\
PReLU & No & Partially \\
\end{tabular} 
\end{table}
\section{Weight initialization}
At the beginning of the training process, assigning the initial values to the weight vector is knows as weight initialization. Rumelhart et al. first tried to assign equal initial values to weight vector but it was observed that the weights  move  in  groups during the weight update and  maintain a symmetry \cite{sodhi2014interval}. This causes the network to fail to be trained properly. As a solution, Rumelhart et al. proposed the   random  weight  initialization  method \cite{mcclelland1987parallel}. The initial weight values are chosen uniformly from a range of [$-\delta,\delta$] \cite{sodhi2014interval}. Xavier  and He normal weight initialization method have been used widely since they have been proposed.

\subsection{Xavier initialization}
Xavier initialization method was introduced in the paper Xavier et al. \cite{unders}. This method proposes that the weights are to be chosen from a  random uniform  distribution bounded between $(-\frac{\sqrt{6}}{\sqrt{n_i + n_j}}, \frac{\sqrt{6}}{\sqrt{n_i + n_j}})$ where $n_i$ is the number of incoming network connections and $n_j$ is the number of outgoing network connections from that layer.  

The initial objective of the Xavier et al. paper was to explore why standard gradient descent from random initialization performs poorly on deep neural networks. They investigate that the logistic sigmoid activation is not suited deep neural networks with random initialization. It leads to saturation of initial layers at very early stage of training. They also find that the saturated units can move out of saturation by themselves  with proper initialization \cite{unders}. 

To find a better range for random initial weight values, they equated the variance of every layer. The variance of the input and output of a layer was equated so that much variance is not lost between the input and output of a single layer. With this idea, the solution found is a range for initial weight values. The range is $$ U [ (-\frac{\sqrt{6}}{\sqrt{n_i + n_j}}, \frac{\sqrt{6}}{\sqrt{n_i + n_j}})] $$ where $n_i$ is the number of incoming network connections, $n_j$ is the number of outgoing network connections from that layer and $U$ is a random uniform distribution.

\subsection{He normal initialization}
He normal weight initialization was introduced in the paper He et al. paper and it has been widely used since it has been proposed \cite{heetal}. This paper addresses two different things - one is a new activation function PReLU which is already discussed in section 2.3.5 and the other one is the new initialization method.

While studying extremely deep networks with rectifier nonlinearities as activation functions, they found if the weights are randomly selected from a range $$N[(-{\frac{\sqrt{6}} {\sqrt{n_i *(1+a^2)}}}, \frac{\sqrt{6}} {\sqrt{n_i *(1+a^2)}})]$$ where $n_i$ is the number of incoming network connections in the layer and $a$ is the parameter of PReLU function and $N$ is normal distribution, the network gets trained fast. This initialization helps to attain the global minima of the objective function more efficiently. The approach of this method is quite similar to the Xavier initialization but it was done in both ways - forward and back-propagation. That is why their method reduces the error faster than Xavier initialization  as they state in their paper \cite{heetal}.  

\section{Discussion}
Xavier et al. strongly suggests that the  sigmoid activation function easily saturates at very early stage of training. This causes the training to fail. It also suggests tanh activation function as a good alternative for the sigmoid because it does not get saturated easily. The main advantage of tanh function over sigmoid, as they show in the paper, is that its mean value is 0 (more precisely, it is a zero-centered activation function) \cite{unders}. Xavier et al. paper, as it states in the section `Theoretical Considerations and a New Normalized Initialization', assumes a linear regime of the network. Sun et al. also claims the same that Xavier initialization typically considers a linear activation function at the layer to bind the gradient values in a range \cite{sun2018improving}. 
\begin{table}[H]
\centering
\caption{Error rates of EM segmentation challenge \cite{ronneberger2015u} }
\label{tab:tableres}
\begin{tabular}{c|c}
Group name  & Error\\
\hline
 \\
*Human value* & 0.000005 \\
\textbf{U-net} & \textbf{0.000353}  \\
DIVE-SCI & 0.000355 \\
IDSIA & 0.000420  \\
DIVE & 0.000430  \\

\end{tabular}
\end{table}

To explore the relation between Xavier and rectifier nonlinearities, He et al. paper discusses an experiment where they experiment a 22 layered neural network and a 30 layered neural network.  ReLU has been used as activation functions and Xavier initialization has been used as weight initializer in the experiment. The experiment shows that the 22 layered network converges while the 30 layered network fails to converge \cite{heetal}. Siddharth et al. shows the reason behind this failure as - `the variance of the inputs to the deeper layers is exponentially smaller
than the variance of the inputs to the shallower layer' \cite{kumar2017weight}. From this, it can be concluded that though Xavier initialization wanted to preserve the variance, but as the network goes deeper, it fails. This is why Xavier initialization works well in the cases of shallow networks. 

He et al. paper states it very clearly that networks with rectifier nonlinearities are easier to train than networks with sigmoidal  activation functions \cite{heetal}. This statement is also supported by papers like Krizhevsky et al. , Glorot et al. \cite{krizhevsky2012imagenet} \cite{glorot2011deep}. This is why He et al. completely ignores the sigmoidal activation functions and focuses on rectifier nonlinearities only and also produces a generalized form of rectifier nonlinearities PReLU. 

The combination of ReLU and He normal initialization has performed well in some of the state of the art papers like Ronneberger et al., Huang et al \cite{ronneberger2015u} \cite{huang2017densely}. In table \ref{tab:tableres} the result of the Ronneberger et al. paper has been shown. It shows that the U-net architecture which uses the combination of ReLU and He normal initialization outperforms the other methods and has the least error rate in a biomedical image segmentation challenge named `EM segmentation challenge'.

\section{Conclusion}
The usage of sigmoidal activation functions is decreasing as the rectifier nonlinearities are being more popular. On one hand, rectifier nonlinearities, especially ReLU, have good performance with He normal initialization in several kinds of networks. On the other hand, the performance of He normal initialization beats the performance of Xavier initialization. Though tanh activation function with Xavier initialization is used but only in cases where the network is not deep. He normal initialization along with rectifier nonlinearities, especially ReLU, get more preference when the network is deep.

\section*{Acknowledgement}
I would like to thank Dr. David Tax, Assistant Professor, Faculty of Electrical Engineering, Mathematics and Computer Science, TU Delft, Netherlands for his supervision and guidance for this survey.
\bibliography{main.bib}{}
\bibliographystyle{unsrt}
\end{document}